# Open-source framework for detecting bias and overfitting for large pathology images

Anders Sildnes[1], Nikita Shvetsov[1], Masoud Tafavvoghi[2], Vi Ngoc-Nha Tran[1], Kajsa Møllersen[2],

Lill-Tove Rasmussen Busund[3,4], Thomas K. Kilvær[5], Lars Ailo Bongo[1]

[1] *Department of Computer Science, UiT The Arctic University of Norway*
[2] *Department of Community Medicine, UiT The Arctic University of Norway*
[3] *Department of Clinical Pathology, University Hospital of North Norway, Tromso, Norway*
[4] *Department of Medical Biology, UiT The Arctic University of Norway, Tromso, Norway*
[5] *Department of Clinical Medicine, UiT The Arctic University of Norway, Tromso, Norway*

**Abstract**—Even foundational models that are trained on datasets with billions of data samples may develop shortcuts that lead to overfitting and bias. Shortcuts are non-relevant patterns in data, such as the background color or color intensity. So, to ensure the robustness of deep learning applications, there is a need for methods to detect and remove such shortcuts. Today's model debugging methods are time consuming since they often require customization to fit for a given model architecture in a specific domain. We propose a generalized, model-agnostic framework to debug deep learning models. We focus on the domain of histopathology, which has very large images that require large models - and therefore large computation resources. It can be run on a workstation with a commodity GPU. We demonstrate that our framework can replicate non-image shortcuts that have been found in previous work for self-supervised learning models, and we also identify possible shortcuts in a foundation model. Our easy to use tests contribute to the development of more reliable, accurate, and generalizable models for WSI analysis. Our framework is available as an open-source tool integrated with the MONAI framework, available at
https://github.com/uit-hdl/feature-inspect.

## I. Introduction

Pathologists examining tissue specimens mounted on glass slides using a high-powered microscope is the gold standard for cancer diagnosis. Currently, glass slides are digitized into whole-slide images (WSI) that comprise billions of pixels and millions of cells. However, it is difficult for humans to extract all relevant features for prognosis in the plethora of information available in a WSI. Deep learning (DL) models therefore show great promise for WSI analysis both by themselves and as decision support for pathologists. For example, DL has demonstrated its usefulness for cancer type classification [3][9], tissue segmentation [19] and analysis of tissue microenvironments [39][41].

An important limitation for WSI model development is the lack of annotated datasets [3]. Consequently, self-supervised learning (SSL) methods trained on larger, unannotated datasets, have recently been used for WSI analysis [6][27]. A popular SSL approach for natural image prediction is contrastive learning (CL) [8][36]. However, many contrastive learning models suffer from a low level of generalizability [29][38]. This means models are overfitted to their datasets and therefore unable to deal with minor fluctuations when applied in a new setting. This is an especially important limitation for medical applications where variability in data, such as differences in staining protocols, scanning equipment, or patient populations, is common. Overfitting in this context can lead to reduced diagnostic accuracy and hinder the adoption of these models in clinical practice. This can also make models biased since they rely on non-clinical features to do their predictions. Lack of generalizability in contrastive learning models may arise from their tendency to rely on *shortcuts*. Shortcuts occur when a model identifies and uses dataset-specific patterns or features, known as artifacts, to make predictions. These artifacts are typically non-generalizable and may be unique to certain datasets. Models trained on WSIs should ideally avoid artifacts since 1) artifacts may lead to overfitting, as they are not representative of broader patterns across datasets, and 2) they lack prognostic or diagnostic relevance, resulting in unpredictable and unreliable model predictions.

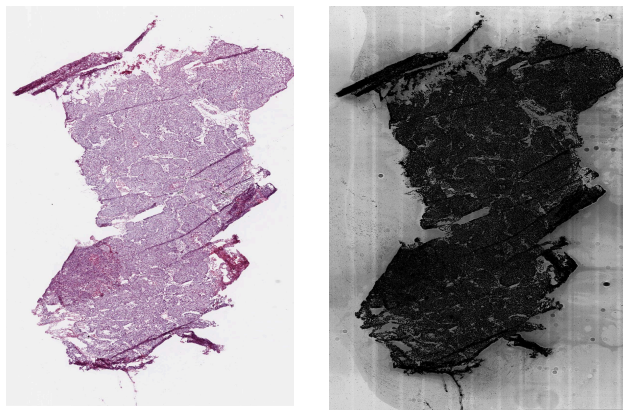

Fig. 1 WSI from TCGA (left) and a modified version (right) contrast enhanced with equalized histograms and stripped of colors. This reveals scanner stripes and air bubbles, not usually visible to humans.

Artifacts can be introduced in tissue preparation, imaging and staining equipment or the chemicals that are used [2][5][22][45]. Fig. 1 shows common artifacts in WSIs. Artifacts can be uniquely present on a single slide or occur in sets processed in batches. Such batch effects are widely known, and accounted for, in other disciplines such as bioinformatics [17], but have only recently gained attention in computational pathology [22][51]. For example, [21] showed that institutions had differing distributions of cancer data, and that models could identify these sites and use site-specific artifacts as shortcuts for predictions. They also show that common color augmentation and normalization techniques were not enough to prevent models from learning these artifacts. [11] did a similar experiment on 8579 slides from The Cancer Genome Atlas (TCGA) datasets and showed that with only a little fine-tuning or re-training a model could map a tile from a slide to the institution where the slide was produced with approximately 80% overall accuracy. Both studies advocate pertinent consideration when sampling and using WSI datasets for deep learning. A part of this consideration is to systematically evaluate model overfitting and data bias. This is a difficult task: models with random parameters will have no bias or overfitting, but poor classification performance. But models with high classification performance may exploit shortcuts and not perform well on unseen datasets.

Recently, several foundational models are becoming available for fields such as histopathology. Foundational models are typically trained with SSL techniques, but on datasets that are orders of magnitude larger than those used for smaller, traditional DL models. While this scale of data can enhance model generalization, it does not eliminate the risk of learning undesirable batch effects. [24] tested nine foundational models trained on WSI data and found that they could all easily detect tissue source sites in TCGA, suggesting that there are considerable batch effects in their training. Choosing the right foundational model for a task can be difficult as there are increasingly many available.

Finding the right test for bias- and overfitting is challenging. It introduces a computational overhead, takes time to implement and interpretation can be subjective. Each model architecture may be different, so testing multiple models will have additional overhead. In related work, it is common to select one or two methods, even though there may be other relevant tests for WSI models. The type of tests include, but are not limited to, heatmaps [28][40][50], vector dimensionality reduction (DR) plots [12][21][51], external validation datasets [10], loss curves [1], linear probing [11][21][24] and estimating domain boundaries [26][42]. While increasing the number of tests could enhance error detection, the need for extra computational resources and/or data (if using external validation) might limit the number of tests that are practical for WSI models. Therefore, we believe there is a need for a general testing framework that scales to large WSI datasets.

Our contributions are threefold. The first is a framework with two key features:

1. **An intuitive framework for detecting and analyzing bias** – We provide user-friendly tools to visualize and assess how batch effects influence model performance and contribute to overfitting in deep learning models. This allows researchers to identify and mitigate potential biases more effectively.
2. **Model-agnostic** methods that can handle the large volumes of data needed for WSI datasets.

The framework uses vector DR plots and linear probing. Both methods are normally used to explore task-specific model classification, but can also be used to explore batch effects in models. These tests work with latent space data, making them model agnostic. This article focuses on histopathology WSIs, but other types of image-data can also be used.

Our second contribution is to demonstrate the framework using two different models. We start by replicating a part of the study of [24] and use the foundational WSI model Phikon-v2 [16] to evaluate tissue-source site (TSS) (also known as acquisition site or institution) biases in TCGA. We believe testing foundational models is especially important since they are known to be susceptible to batch effects [24]. We also test a model that we train on only a single dataset. We chose MoCo V1 [20], a CL SSL model with Inception-V4[43] as the encoder and decoder. MoCo V1 was also used by [51] to detect slide-level biases, making it a relevant choice for detecting and understanding potential batch effects in our study.

The third contribution is making the framework efficient and scalable for consumer-level GPUs. We do this by utilizing code optimizations implemented in the open-source Medical Open Network for Artificial Intelligence (MONAI) [4] framework. MONAI wraps PyTorch [14] models with extensions that make the framework appropriate for the scale and testing needed for medical data such as WSIs. Furthermore, since MoCo V1, and many other SSL models consume a lot of VRAM, we also demonstrate the use of sequential checkpointing to reduce VRAM usage. Such optimizations allow debugging on consumer-level hardware, which removes the need for expensive computers to inspect models.

Our framework addresses core challenges of developing robust and unbiased deep learning models for WSI analysis. By optimizing and standardizing model training processes and doing systematic evaluation, we facilitate generation of more reliable, accurate, and generalizable models for WSI analysis. The framework is open source under the Apache 2.0 license and available at https://github.com/uit-hdl/feature-inspect.

II. FRAMEWORK FOR INSPECTING FEATURE-VECTORS

In this section, we describe the design and implementation of our proposed framework to inspect model features from DL models. We then show an example workflow using the framework for bias and overfit-detection on a WSI dataset.

## A. Architecture and Design

The architecture for our suggested framework (Fig. 2) is designed to work on a single-GPU consumer-grade computer. It can be used during DL model training or to evaluate a pre-trained model. Our framework is exposed as a python API which is invoked with a few lines of code. The requirements for model evaluation is that the user has a trained model available, a dataset and labels - which could either be clinically relevant, such as tumor prognosis or to investigate biases, such as labelling data with their TSS.

To start testing, users first extract features using inference with a DL model (Fig. 2a). The features are embedded representations of the input. For SSL models these have around 1000 elements with floating point numbers. The output from inference is used as input to the bias/overfit detection module (Fig. 2b). The user also supplies a set of either sensitive or prognostic variables for each feature vector. For example, to explore how biased the model is towards gender, the user can associate images/feature vectors with a gender label. The output from the bias/overfit detection is either printed to a text file, or for UMAPs, a website for visual inspection. After inspecting the results, a model developer can then consider using another model, or implementing different techniques such as regularization, data cleaning or data weighing to make the model more robust to unseen data.

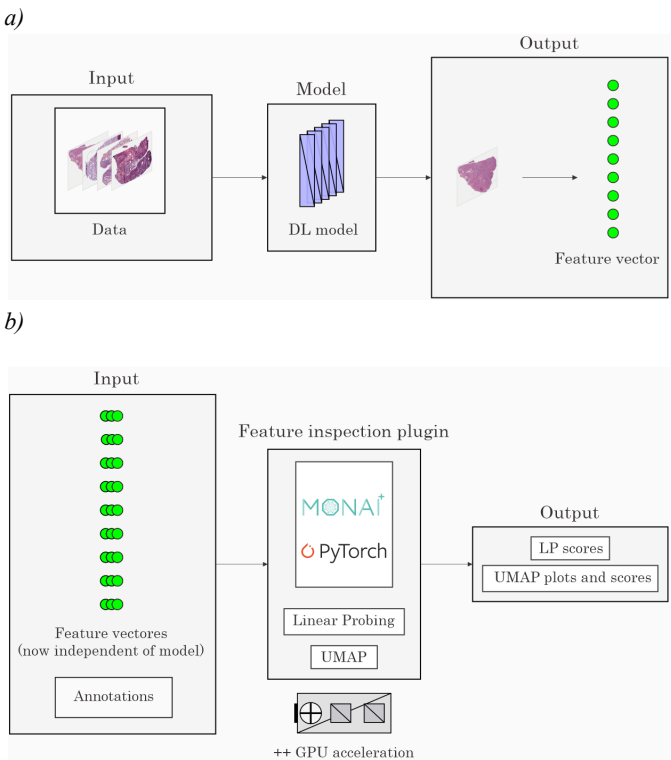

Fig. 2 a) Process of extracting feature vectors from a trained model. The output consists of arrays of floating-point numbers that represent the underlying features captured by the model. b) Feature-inspect pipeline. For a set of feature vectors, the framework can be inspected using UMAPs, linear probing or both, using LP as a scoring function for UMAPS.

For our linear probe and model training, we use MONAI. It is a widely used PyTorch extension for medical data. It provides tools for image preprocessing, augmentation, and analysis. MONAI also provides optimizations such as distributed processing of models and data, smart caching and improved image processors (such as cuCIM from [46]). Additionally, MONAI makes it easy to bundle the code with additional tools such as TensorBoard[1], which makes it easy to track model loss, validation accuracy and similar metrics in a web interface. MONAI also has other built in interpretability tools such as heatmaps and occlusion maps that can be used in addition to the tools we provide in our framework.

## B. Bias detection workflow - a demonstration

To show our framework's use, we provide a case study for finding potential biases and overfitting in a DL model. A minimal workflow is detailed in Listing 1. First, the training data and annotations are loaded from a specified directory (step 01). This data is WSI-tiles in JPG format with slide-level labels for which institution the WSIs originated from. Feature vectors are then extracted after a forward pass of the model. Steps 03 and 04 produce UMAP plots and linear probing results. The results can either be printed to console, rendered to a TensorBoard, or written to HTML. Listing 2 has a MONAI training loop for a simple, supervised model. To avoid slowing down the training loop too much, the UMAP and linear probing can be configured to only run for a given epoch interval. This allows users to visualize the training process over time.

```
# as per Fig. 2a)
01 data, labels = load_data("data_path")
02 features = model.inference(data)
# as per Fig. 2b)
03 umap_inspect.make_umap(features, labels)
04 linear_probe.linear_probe(features, labels)
```

Listing 1: general workflow of our framework to do inspection and analysis of models trained on a given dataset.

```
evaluator = SupervisedEvaluator(
  val_data_loader=dl_val,
  network=model,
  val_handlers=[
    feature_inspect.UmapExplorer(every_n=20),
    feature_inspect.LinearProber(every_n=20)])
trainer = SupervisedTrainer(
  max_epochs=epochs,
  ValidationHandler(1, evaluator),
  train_data_loader=dl_train,
  network=model,
  inferer=SimpleInferer())
trainer.run()
```

Listing 2: outline for how to attach the feature-inspect tool to a classical MONAI model training loop. The code in bold is the additional lines of code needed, the rest are common configuration options for the MONAI training setup. "dl" is short for "dataloader". For more information, see MONAI documentation at https://docs.monai.io/en/stable/engines.html.

---

[1] A tool to view plots, texts, figures and more: https://www.tensorflow.org/tensorboard

## C. Overfit- and Bias Detection Methods

*1) UMAP plots for qualitative analysis*

Uniform Manifold Approximation and Projection for Dimension Reduction (UMAP) [32] plots scale well to large inputs and offer a visually appealing interface for data exploration aiming to find data anomalies, wanted- or unwanted clusterings, and more. UMAP plots have become common to show that a model is more fair or less biased [29][30], also in histopathology [12][15][21][28][39][51].

UMAP computations are generally fast, although for millions of data-samples, CPU-based implementations can take several minutes. A UMAP is made by constructing a graph of high-dimensional data, then using a stochastic gradient descent to minimize the difference between the high- and low-dimensional representation. The original UMAP algorithm is executed on CPU(s) [31], but several libraries offer GPU-optimized implementations. [34] have published a GPU-based implementation in the "cuml" package [37]. Their speedups for GPU compared to a naïve CPU-implementation was up to 100x, depending on data and hyper-parameters. We use this as the default option for computing UMAPs in our framework.

We have built a web interface that allows a user to quickly navigate through UMAP plots (Fig. 3). The web interface makes it easy to share results and enable data exploration. We embed UMAP plots with different parameters together and allow side-by-side viewing with embeddings from raw data. Plots from raw data can help to unveil whether the model is doing a good clustering or if it is merely picking up obvious patterns such as large shapes or color intensities. We also provide different scoring functions that can be used to assess the quality of a UMAP plot. Silhouette scores [54], a metric for clustering quality, is commonly used. We also provide "k-nearest neighbour" (KNN) and Spearman correlation (CPD) from [24], metrics to measure the quality of local versus global structure preserved in the plot from the original high-dimensional embeddings.

For a thorough analysis, several different UMAPs should be made. There is no guideline for how many points are needed to debug a model. Using few data samples can enhance interpretability by making patterns easier to observe, but these patterns may also arise from the choice of data that is used. Using many points can reveal global patterns, but also make smaller anomalies hidden as noise. Doing multiple plots with different sample sizes is therefore important for debugging.

*2) Linear probing for quantitative analysis*

For UMAP interpretation, we build a scoring function inspired by [11], [21] and [24]. They used linear probing (LP) (Fig. 4) to identify whether a model was able to predict tissue-source-site from WSIs in DL models. Tissue source site is not visible in an image, so a model using it for prognosis or diagnosis may be biased if the majority of patients from that TSS have similar diseases.

To do linear probing, [11] used models that were pre-trained on either WSIs or the ImageNet database [13]. They froze all the weights of the model, but added new layers to the end of the neural network. Freezing all the weights and attaching layers is equivalent to extracting the feature-vectors and training a separate linear model. Thus, this technique is model agnostic. The attached linear layers were trained for up to five epochs, with the same input, but now using TSS as a label. The final output layer would therefore have *n* neurons, where *n* was the number of TSSs. Their accuracy was overall high, ranging from around 60 to 80% accuracy for datasets from TCGA. This scoring function is therefore similar to a UMAP - in the sense that we reduce from high to low dimensions - but it only provides a single number, accuracy, as an output.

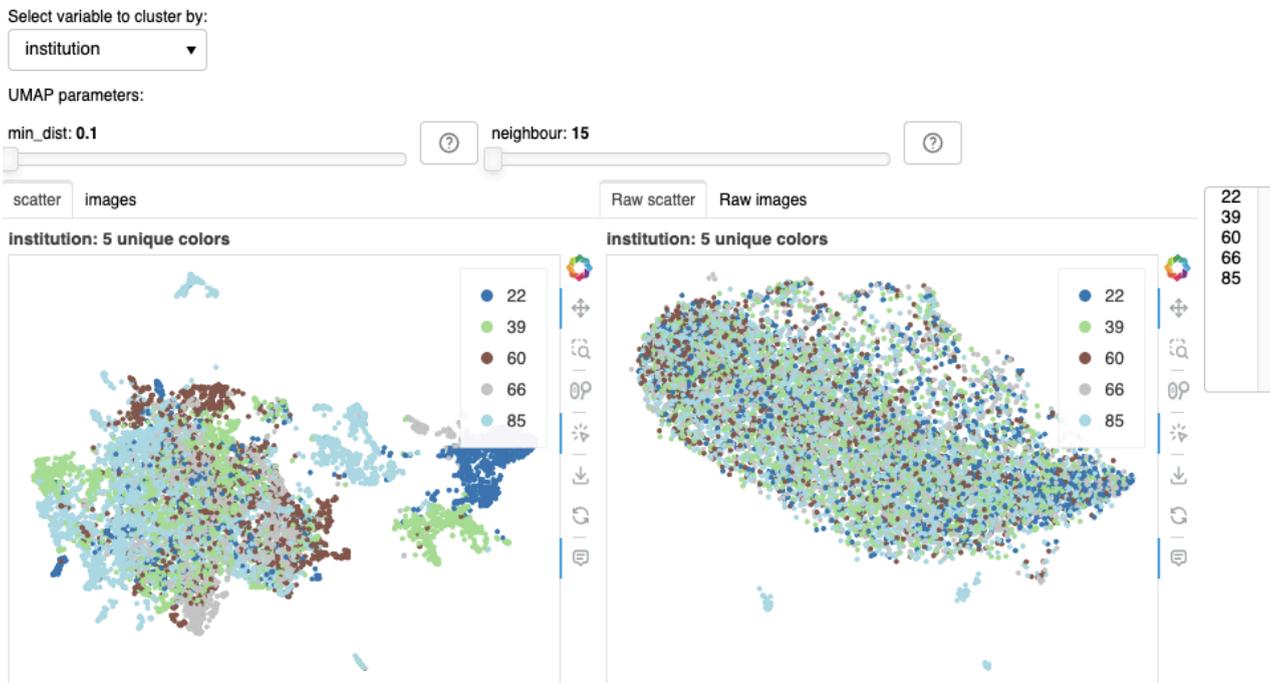

Fig. 3 Example UMAP plot from our web user interface. The top-left drop-down menu allows the user to select a variable to cluster by. Below are two common UMAP parameters that the users can drag and select to see different UMAPs below. The plots come with common interaction tools such as zooming and selecting regions. On the right, there is a list to hide or show specific classes. The left UMAP plot shows a UMAP from feature-vectors and the right shows a UMAP plot with the same data without using any model to extract embeddings.

Linear probing is quick to compute and intuitive to interpret. Overall memory usage is low since there is no intermediary information stored in the LP model. The main constraint is how many samples to use and the size of each feature-vector. A typical batch size of 256 only used approximately 1 GB of VRAM which can be accommodated on most computers.

Similar to UMAPs, there is no best practice for choosing data for linear probes. Limited samples per class can affect LP scores. [11] split TCGA into two groups and trained two-layer linear probes: one on institutions with the most tiles (76% TSS accuracy) and the other on those with fewer slides (56% accuracy). These results suggest that LP performance is influenced by the number of samples per class. We therefore use LP for potentially bias-inducing variables that have many samples per class (such as TSS). For other potential biases, such as slide-level biases investigated by [51], there may not be enough labels to thoroughly investigate LP accuracy.

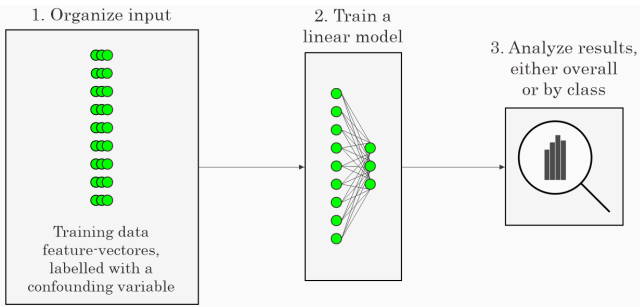

Fig. 4 Linear probing. Feature-vectors from inference are labelled and then trained with a linear layer. The linear layer has maps from *n* to *m* neurons, where *n* is the number of features from the model, and *m* is the number of classes to predict. The result will be the output neuron with the highest score, which can be used to measure prediction accuracy.

III. METHODOLOGY - FRAMEWORK VALIDATION AND PERFORMANCE OPTIMIZATION ON CONSUMER-LEVEL GPUS

This section presents the use case and the performance optimization techniques applied in our framework.

*A. Use case: TSS-level bias in TCGA*

[11][21][24] all investigate tissue source site (TSS) -level bias in models trained on the datasets from The Cancer Genome Atlas [47]. TSS biases can be harmful, since the model uses this shortcut to identify sources with more disease than others. The model performance may therefore not transfer to new, unseen clinics.

From TCGA, we use the Lung Squamous Cell Carcinoma (LUSC) WSI dataset. The LUSC dataset contains 1100 tissue slides from 495 patients. Of these, 753 slides have tumors and 347 are normal. All the slides were scanned by the same type of Aperio scanner. The tissue slides are scanned at either 20x or 40x magnification.

Similar to [24], we primarily look at the top five TSSs when probing for TSS bias. This is to ensure that we have enough samples per class to detect TSS-level biases. If we used TSS with only a few samples, it would be harder to detect bias patterns and especially to train a LP classifier that depends on having multiple samples to adjust the model weights and parameters. To choose the top five TSSs, we assign a rank after creating tiles from slides and filtering for whitespace.

*B. SSL model trained on a single dataset*

To investigate TSS-biases in TCGA, we train a CL model using a similar architecture as in [51] where they investigated slide-level biases in CL models. They used a MoCo v1 model [20] with InceptionV4 [43] as encoder and decoder. The InceptionV4 encoder is used for downstream classification tasks. The output layer has 1536 neurons. Each WSI is preprocessed by tessellating into tiles of size 1024x1024 with 25% overlap. Tiles with more than 85 % whitespace are removed. Then each tile is downsampled to 299 x 299 pixels and color normalized using the Vahadane method [48] using a reference image from TCGA, chosen by [51], because it had a clear stain well defined cell structure. The reference image is available from both in [25], and on our github repository: https://github.com/uit-hdl/code-overfit-detection-framework.

For training MoCo V1, we use a 70-15-15 split. The training was done on all the slides available in TCGA-LUSC, not just the top five TSSs. The number of tiles per split is given in Table 1.

| *Split set* | Amount of samples in % | Number of tiles | Number of tiles after balancing for top 5 TSS |
|---|---|---|---|
| train | 70 | 189322 | 67655 |
| validation | 15 | 23666 | 14497 |
| test | 15 | 50621 | 14498 |
| total | 100 | 264609 | 96650 |

Table 1: Data splits used for training on TCGA. Slides from one patient can only occur in one split. This can result in a different number of tiles for each train/validation/test stratification. The numbers are therefore averaged from three runs.

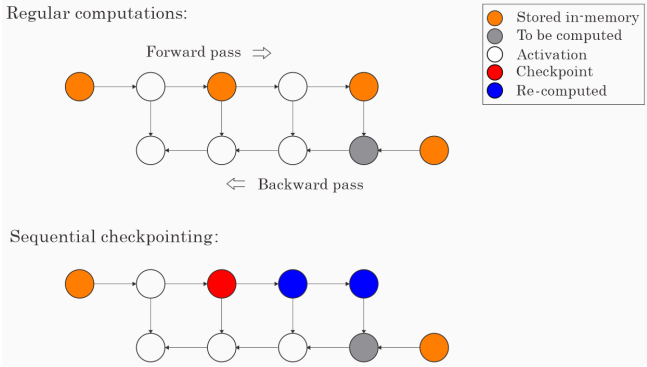

Fig. 5 an example of sequential checkpointing. In the regular computation (above), information from the forward pass of the model is stored in memory. To do a backwards pass, this information is passed on for calculations. For sequential checkpointing (lower part), not all information is kept in-memory. Therefore, to get the information needed for the gray node, the latest checkpoint is found (red), and information from that checkpoint is re-calculated (blue nodes).

*C. Reducing memory usage: sequential checkpointing*

A significant challenge for training an SSL model is having enough computational resources. One of the primary constraints is the amount of VRAM. Today's consumer-level computer hardware has a single GPU with up to 24 GB of VRAM. This is not enough to load several foundational or SSL models and therefore hinder consumers from fine-tuning, inspecting or re-training a model. For our MoCo model, we are only able to train with a batch size of 64 on a GPU with 32 GB of VRAM.

To reduce GPU memory usage in MoCo V1, we implement sequential checkpointing [14]. It divides the forward pass results into checkpoints, reducing memory usage. Conventional training stores all intermediate values for fast backpropagation, but consumes $O(n)$ memory for a network of depth $n$. In contrast, checkpointing reduces memory consumption by only storing information in some nodes, but it increases computational overhead to a quadratic run-time due to full forward-pass recalculations. Fig. 5 illustrates regular training and training with checkpoints. As in [7], we set the number of checkpoints to be $\sqrt{n}$. So for MoCo V1, we use four checkpoints, which allows us to store a batch size of 128 tiles in 10 GB of VRAM, a reduction by more than 50%.

*D. Phikon-v2 foundational model for histopathology*

For TSS-level evaluation, we select Phikon-v2 [16], a DinoV2-based model with 1 billion parameters [36]. It is trained on 460 million pathology tiles from public datasets, including TCGA. Phikon-v2 therefore has a large size and extensive training set. This makes it ideal for illustrating batch effect detection, and to evaluate the scalability of our framework.

*E. UMAPs to detect TSS*

To detect TSS-biases using UMAP plots, we use two datasets of randomly sampled tiles. The first has 10 000 tiles to detect local patterns and the second has 100 000 tiles to detect global patterns. The tiles are labelled by their TSS. We use default hyper-parameters for the UMAP plot, which by the implementation from the authors [31] has "min_dist" 0.1 and "n_neighbors" as 15.

*F. Training an linear probe for TSS classification*

To evaluate if models can identify tissue source sites from tiles only, we train a linear probe from feature vectors extracted from our Phikon-v2 and InceptionV4 encoder. The split configuration for training the linear probe on TCGA-LUSC are in Table 1. We trained the LP for 20 epochs using the Adam optimizer [53]. Accuracy is averaged from three runs using different training, validation and test splits.

*H. System Environment*

We run our code on a consumer-level PC with an Intel i9-11900K processor, one or two NVIDIA Tesla V100 with 32 GB of VRAM, and main memory at 128 GB of RAM. We use PyTorch version 1.13.1, MONAI version 1.3.0 and CUDA version 12.2. The number of workers for the training loop was 6. For each model trained to do classifications we have collected averages from three runs using different training, validation and test splits.

## IV. RESULTS

This section presents results from training a SSL model and identifying TSS-level biases in Phikon-v2 and InceptionV4. We also present the computational performance for both LP, Phikon-v2 and sequential checkpointing.

*A. TSS classification on TCGA-LUSC*
*1) Linear probing TSS detection accuracy in InceptionV4 vs. pre-trained Phikon-v2:*

For Phikon-v2, we trained a linear probe to map tiles to TSS. The accuracy for mapping tiles to TSS was 63.2%. [24] also used LP with Phikon-v2 on TCGA-LUSC, although they used different TSSs and used different color normalizations on their tiles. Their accuracy numbers were between 90% for non-normalized tiles and 70% for normalized tiles. This seems to indicate that color normalization does play an impact, although it is not sufficient to remove TSS artifacts.

The LP trained on the same tiles extracted from InceptionV4 got an accuracy of 34%, only 14% better than random guessing. This is close to half of Phikon-v2, which indicates that a self-trained model may have significantly less batch-effects.

*2) UMAP plots from Phikon-v2 and InceptionV4 for TSS clustering:*

To identify TSS-level biases from a UMAP plot, we primarily look for regions/clusters of tiles that do not overlap when coloring for TSS. This indicates that a model is mapping tiles from the same TSS close in embedded space, relying on TSS artifacts. Fig. 6 shows UMAP plot embeddings from Phikon-v2 and InceptionV4. Each UMAP used the same tiles randomly sampled from the top five tissue-source sites in

TCGA-LUSC. We used both 10 000 and 100 000 samples to detect local and global patterns. In both examples, Phikon-v2 has the least amount of TSS overlap, particularly for dark blue (22), light blue (85), and green (39) tiles. This suggests that the model is aware of- and uses TSS artifacts. It also aligns with high TSS LP scores. The InceptionV4 encoder from MoCo V1 shows high TSS overlap. This suggests that the model is not aware or uses TSS artifacts. It is difficult to compare the plots from InceptionV4 and Phikon-v2, since the morphology differs and there are different numbers of clusters. A model developer would have to consider either using a different set of points or reducing the number of points to get a more clear interpretation. For the InceptionV4 plot with 100 000 points, the points overlap so much that it seems like institution 85 "dominates" the input. This illustrates the benefit of our web view, which allows the user to filter away 85 to inspect the other tissue source sites. Though, this does not reveal any other, obvious patterns for estimating TSS bias, suggesting the model is not aware of TSS artifacts. This also aligns with the lower LP scores.

Overall, Phikon-v2 seems to be more aware of TSS artifacts than InceptionV4 models. We believe it is because Phikon-v2 has been trained on much larger datasets: with more data, the contrastive learning component is less likely to have multiple tiles from the same TSS per batch. If each TSS occurs alone in a batch, and has a strong site-specific artifact, the model can use these artifacts ("shortcuts") to compare/contrast tiles. The second reason may be that Phikon-v2 has many more parameters, which allows it to learn more fine-grained signatures that may include TSS-level artifacts. More research is needed to fully understand how to prevent learning shortcuts and their impact on model accuracy.

B. *Computational performance*

*1) CL model training execution time:*

We investigated the impact of sequential checkpointing and batch size on VRAM usage, training time, and model performance. We measured VRAM usage once every second while training the model in a separate thread using a package called "pynvml" version 12.0.0 [55]. First, we observed that implementing four checkpoints in MoCo V1 reduced VRAM usage by approximately 50%. However, this came at the cost of increased execution time. To quantify this, we trained the MoCo V1 model with a batch size of 128, both with and without sequential checkpointing, using a training dataset of 189,322 tiles (similar to Table 1). Without checkpointing, a single training iteration required approximately three days. Adding sequential checkpoints increased the training time by 29%, extending the total duration by nearly a day for the TCGA-LUSC dataset. This highlights the trade-off between memory efficiency and execution speed.

To explore the effect of batch size, we conducted a second experiment using a batch size of 256, which required approximately 20 GB of VRAM. This configuration decreased execution time by 26.3%. We attribute this improvement to the more efficient utilization of the GPU, likely due to reduced idle time as data was loaded by each worker.

In our third experiment, we introduced a second GPU to the setup using PyTorch's "DataParallel" [56]. Without sequential checkpoints, training time decreased by approximately 25% compared to the standard single-GPU training loop. However, when sequential checkpoints were enabled with a batch size of 256, the training time remained comparable to that of a single GPU, suggesting no additional benefit from the second GPU. This could be due to various factors, such as time spent on gradient synchronization, additional overhead of re-computing nodes for each GPU or IO overhead. However it is notable that sequential checkpointing did not incur a penalty in our setup.

Finally, using a batch size of 256 had negligible effects on downstream tasks, including linear probes and UMAP plots. The UMAP visualizations for different configurations were qualitatively similar, and the linear probe accuracy for TSS- remained within 5% of baseline performance.

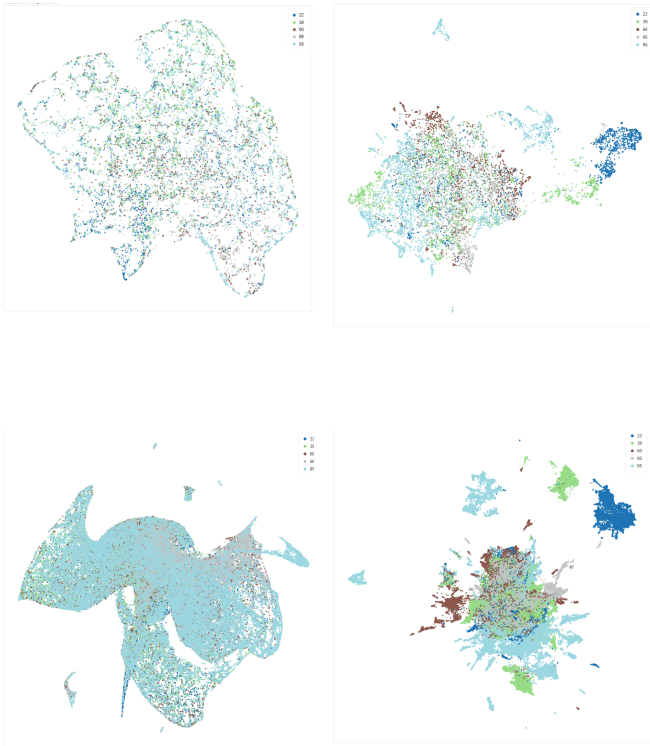

Fig. 6 UMAP embedding from InceptionV4 (left) and Phikon-v2 (right). The top row uses tiles randomly sampled from 10 000 of the top five TSSs. The bottom row uses 100 000 points. The left and the right side use the same points for UMAP clustering.

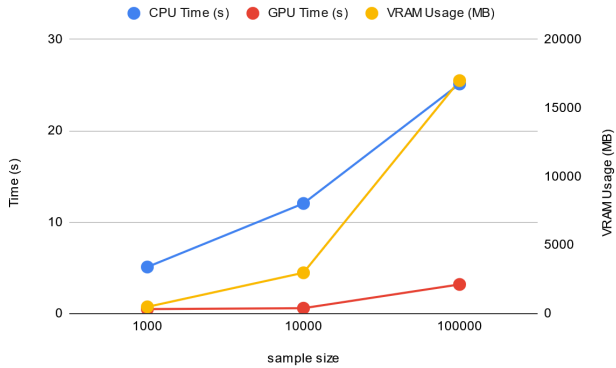

Fig. 7 a comparison of speed for UMAP on CPU vs GPU and VRAM usage for 1000, 10000 and 100000 points. While the speedup for 100 000 points is better on the GPU, it also requires more VRAM.

*2) UMAP plot GPU acceleration:*

Our framework is intended to be used by consumer-level hardware. We compare existing implementations of the UMAP algorithm on both CPU and GPU. We do not only focus on time usage but also VRAM. Fig. 7 has a comparison of CPU and GPU execution times together with VRAM usage. For speedup, we find similar results to the authors of the GPU-implementation in [34]: a speedup of up to 5x with 1000 samples and 8x for 10000 samples.

The VRAM usage for processing 100,000 feature vectors extracted from tiles by InceptionV4 and Phikon-v2 is approximately 17 GB. Such a requirement may pose challenges when analyzing large datasets, such as the 460 million tiles used to train Phikon-v2. While 100,000 samples may suffice for certain analyses, optimal usage would benefit from generating multiple plots under varying sampling conditions and classes. For significantly larger datasets, users would need access to a more powerful GPU setup or rely on CPU-based implementations, which could result in rendering times up to 100 times slower.

*3) Computational resources for linear probing:*

Since linear probing only uses a single linear layer, it is cheap to execute and does not require extensive computations. We trained our models for 20 epochs. The accuracy for TSS predictions generally did not improve after 20 epochs, although this could be impacted by learning rate, choice of optimizer and other hyper-parameters. [11] used five epochs, but also had a much bigger dataset sampled from 8,579 slides. We have not found any literature that discusses how many epochs are needed to achieve a *good enough* accuracy. Nevertheless, in our experiment each epoch of linear probing was cheap to add - for 67655 tiles converted to 1536-element feature vectors in batches of 256 elements, each epoch added 0.34 seconds. The total time used, with validation, logging and saving intermediate states is about 30 seconds. This duration is brief enough to allow users to explore multiple options and can be added to existing model training routines without significant delay.

## V. DISCUSSION

In this work, we developed an open-source, model-agnostic framework for detecting bias and overfitting in deep learning models for whole-slide image (WSI) analysis. Our framework provides an intuitive and computationally efficient approach, integrating UMAP visualizations and linear probing to systematically evaluate model robustness. This contribution was necessary because existing bias detection methods are often model-specific, computationally expensive, and require extensive customization, making them impractical for large-scale WSI datasets. By enabling rapid identification of batch effects, particularly tissue source site (TSS) biases, our framework improves the reliability and fairness of deep learning models in histopathology. Its scalability to consumer-grade GPUs ensures broader accessibility. However, users should be mindful of limitations, including biases introduced by data subsampling and hyperparameter choices.

We focused on TSS-level biases in TCGA but recognize that other sources of bias, such as slide-level batch effects [51], warrant further investigation. Class imbalance presents a significant challenge when testing for additional variables, particularly for underrepresented classes like unique slides. Addressing this may require conducting more tests with varied data splits to ensure robust and meaningful results. Moreover, hidden confounders in the data remain a concern. Even with low LP scores and UMAP plots showing no clustering, one cannot definitively conclude that a model is unbiased; it merely indicates the model has been checked for specific confounders, such as TSS.

The TSS-artifacts used by foundational models underscores a critical challenge in model evaluation: balancing the trade-off between high accuracy and potential bias. Similar dilemmas are observed in other areas of deep learning, such as adversarial robustness, where models are trained on noisy images to enhance generalisability [49], often at the cost of prediction accuracy.

Interpreting UMAP plots poses challenges due to information loss during dimensionality reduction, non-uniform feature representations, and parameter sensitivity. While interpretation depends on the viewer, UMAP plots are scalable to large datasets and provide an intuitive interface for exploring data anomalies and clustering patterns.

We demonstrate that sequential checkpointing not only enables the use of larger models but also accelerates training by allowing increased batch sizes. This improvement makes it practical to use SSL models for training and fine-tuning on consumer-grade hardware, broadening their accessibility. Our improvements may be specific to MOCO V1 and our hardware. However, our results encourage more exploration for other developers to find optimal settings to run models on memory-constrained computers.

Evaluating UMAP plots and LP requires converting image data into feature-vectors through model inference, which can be time-intensive with large sample sizes, potentially taking several minutes. To address this problem, we optimize runtime by integrating LP and UMAP evaluations into the training process and caching feature vectors. To speed up inference for pre-trained models, one can either use compressed models or compress feature-vectors to reduce the data size - although further research is needed to investigate the impact on UMAP plot and LP-score interpretations.

VI. CONCLUSION

We introduced an open-source framework for detecting bias and overfitting in deep learning models for whole-slide image (WSI) analysis. By leveraging UMAP visualizations and linear probing, our framework enables efficient detection of batch effects, particularly at the tissue source site (TSS) level. Our results demonstrate that even large-scale foundation models, such as Phikon-v2, exhibit TSS artifacts that impact generalizability, whereas contrastive learning models trained on a single dataset show reduced bias. However, it could still be that the foundational model has better performance for other downstream tasks. Balancing model robustness and performance in computational pathology is a challenge.

Additionally, we optimized the framework for consumer-grade GPUs using sequential checkpointing, making large-scale model debugging more accessible. By integrating systematic bias detection into model evaluation workflows, our framework helps researchers and clinicians develop more reliable and generalizable WSI models. Future work should explore additional sources of bias, such as slide-level artifacts, and refine techniques to mitigate shortcut learning while maintaining high diagnostic accuracy.

VII. ACKNOWLEDGEMENTS